\documentclass[letterpaper,10 pt,conference]{ieeeconf}    % Comment this line out if you need a4paper

\IEEEoverridecommandlockouts                              % This command is only needed if 
\usepackage{graphics} % for pdf, bitmapped graphics files
\usepackage{xcolor}
\usepackage{amsmath} % assumes amsmath package installed
\usepackage{amssymb}  % assumes amsmath package installed

\usepackage[caption=false,font=normalsize,labelfont=sf,textfont=sf]{subfig}
\usepackage{textcomp}
\usepackage{stfloats}
\usepackage{url}
\usepackage{verbatim}
\usepackage{graphicx}
\usepackage{threeparttable}
\usepackage{booktabs}
\usepackage{wasysym} % for \diameter
\usepackage{siunitx} % for SI units
\usepackage{booktabs}
\usepackage{soul}

\newcommand{\revision}[1]{\textcolor{black}{#1}}
\title{\LARGE \bf
\revision{Fabrication and Characterization of Additively Manufactured Stretchable Strain Sensors Towards the Shape Sensing of \\Continuum Robots}}
\author{Daniel C. Moyer, Wenpeng Wang$^*$, Logan S. Karschner, Loris Fichera,
and Pratap M. Rao% <-this % stops a space
\thanks{This material is based upon work partially supported by the National Science Foundation under Grant numbers DGE-1922761 and 2341532. Any opinions, findings, and conclusions or recommendations expressed in this material are those of the author(s) and do not necessarily reflect the views of the National Science Foundation.
The first two authors contributed equally to this work.
\textit{The asterisk indicates the corresponding author.}
}% <-this % stops a space
\thanks{D.C. Moyer, L.S. Karschner and P.M. Rao are with the 
Department of Mechanical and Materials Engineering, 
Worcester Polytechnic Institute, Worcester, MA 01609, USA.}%
\thanks{$^*$W. Wang and L. Fichera are with the Department of Robotics Engineering, Worcester Polytechnic Institute, Worcester, MA 01609, USA (e-mail: {\tt\small wwang11@wpi.edu})}%
}
\begin{document}
\maketitle
\thispagestyle{empty}
\pagestyle{empty}
%%%%%%%%%%%%%%%%%%%%%%%%%%%%%%%%%%%%%%%%%%%%%%%%%%%%%%%%%%%%%%%%%%%%%%%%%%%%%%%%
\begin{abstract}
This letter describes the manufacturing and experimental characterization of
novel stretchable strain sensors for 
continuum robots.
The overarching goal of this research is to provide a new solution for the
shape sensing of these devices.
The sensors are fabricated via direct ink writing,
an extrusion-based additive manufacturing 
technique.
Electrically conductive material (i.e., the \textit{ink}) is
printed into traces whose electrical resistance
varies in response to mechanical deformation.
The principle of operation of stretchable strain sensors is analogous to that of conventional strain gauges,
but with a significantly larger operational window thanks to their ability to withstand
larger strain.
Among the different conductive materials considered for this study,
we opted to fabricate the sensors with a high-viscosity
eutectic Gallium-Indium ink, which in initial testing 
exhibited high linearity ($R^2 \approx$  0.99), gauge factor $\approx$ 1, 
and negligible drift.
Benefits of the proposed sensors include (\textit{i}) ease of fabrication, as they can be conveniently printed in a 
matter of minutes;
(\textit{ii}) ease of installation,
as they can simply be glued to the outside body of a robot;
and (\textit{iii}) ease of miniaturization, which enables integration into
millimiter-sized contnuum robots.
\end{abstract}

%% Main Sections
\section{Introduction}
Continuum Robots are a class of flexible, slender manipulators
that can bend and twist continuously along their length, allowing
them to navigate through complex and tortuous spaces.
Research interest in these robots has increased
tremendously over the last fifteen years, as documented
in a recent survey by
Russo \textit{et al.}~\cite{Russo2023}.
Continuum robots hold considerable potential in numerous
applications, including interventional
medicine~\cite{Dupont2022,Burgner2015},
industrial inspection and repair~\cite{Qi2024,Russo2021},
and underground exploration~\cite{Coad2019}.
Shape sensing is crucial for the control
and navigation of continuum robots.
By monitoring their shape, continuum
robots can perform tasks with high accuracy,
avoid obstacles, and have safe interactions with humans.
A comprehensive survey (up to 2016) of shape sensing
methods for continuum robots can be found in the 
review by Shi and colleagues~\cite{Shi2016}.
Existing approaches include, among others, methods based on computer vision~\cite{Shentu2024,Ferguson2024,Gao2022,Zhang2022,Zhao2022RA-L,Su2016}, 
electromagnetic tracking~\cite{Lilge2022,Song2021,Mahoney2016},
and Fiber-Bragg
Grating~\cite{Chitalia2020,Sefati2018,Xu2016,Ryu2014}.
It is also possible to implement shape sensing by 
integrating traditional strain gauges \revision{(i.e., strain gauges that consist of a metal foil bonded onto a flexible plastic sheet)} into the body of a continuum 
robot, e.g.,~\cite{Zhao2022}.
This solution is appealing because strain gauges
are relatively inexpensive and
straightforward to use.
However, the body of a continuum robot can experience
large deformations that exceed the operational
\revision{strain} limits of traditional strain gauges \revision{ (typically less than 1\% \cite{Zhao2025}). Therefore, measures need to be
taken to prevent the sensors from breaking, 
such as installing them into semi-rigid enclosures
(as in~\cite{Zhao2022}), which adds to the 
complexity of their integration and use.}
Furthermore, it can be a challenge to install 
strain gauges in the body of some continuum robots,
particularly those developed for surgical applications,
due to their minuscule size.
To overcome these challenges, in this letter,
we propose to explore the use of resistive \textit{stretchable strain
sensors}.
\begin{figure}
    \centering
    \includegraphics[width=1\linewidth]{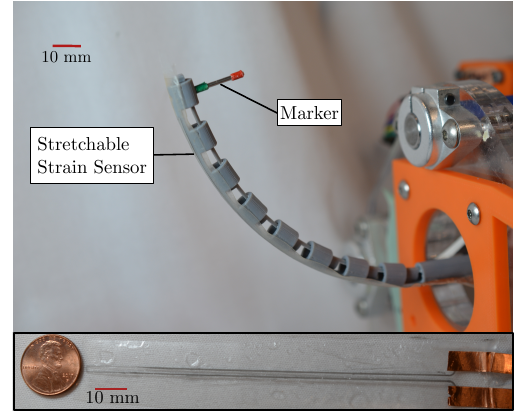}
    \caption{\revision{(Top): }Stretchable strain sensor installed onto a concentric push-pull robot~\cite{Oliver2021}.
    \revision{(Bottom): Sensor specimen.
    The strain sensors we describe in this
    letter can withstand much larger strains than
    traditional strain gauges and are therefore
    suitable for installation on the exterior 
    surface of a continuum robot.}
    }
    \label{fig:fig-1}
\end{figure}
This is a class of sensing devices 
that consist of a
thin and highly stretchable
electrically conductive material~\cite{Ma2022,Liu2021}.
The principle of operation is analogous to that of a
conventional strain gauge (i.e., deformations of the
sensor produce a change in electrical resistance), 
but with a larger operational window.
Within robotics, these sensors are currently being
explored for applications in soft prosthetics and
wearable devices, as well as the monitoring 
of soft robotic grippers and
actuators~\cite{Souri2020,Abbara2023,Koivikko2018}. 
In this letter, we show that these sensors are also suitable for 
estimating the bending of continuum manipulators (see Fig.~\ref{fig:fig-1}), thus 
providing a new option for the shape sensing of these devices.
Throughout the remainder of this letter,
we describe the design, fabrication, and
experimental characterization of novel stretchable 
strain sensors for continuum robots.
The sensors are manufactured via direct ink
writing, a rapid prototyping method in which
the conductive material (i.e., the \textit{ink})
is additively printed into a prescribed shape,
similarly to the process used in extrusion-based
3D printers~\cite{Liu2021}.
We first report the results of a study aimed at
identifying a suitable conductive ink 
out of a set of three candidate
materials, 
namely a silver-elastomer composite, 
a carbon-elastomer composite, and a gallium-indium-based 
liquid metal
composite.
We then introduce stretchable strain sensors
for two types of continuum manipulators,
i.e.,
a push-pull concentric tube robot~\cite{Childs2024}
(\diameter~8 mm), and a notched-tube continuum
wrist~\cite{Pacheco2021} (\diameter~1.1 mm).

\section{Sensor Fabrication and Evaluation of the Conductive Inks}
\label{sec:methods}
\begin{figure}
    \centering
    \includegraphics[width=\linewidth]{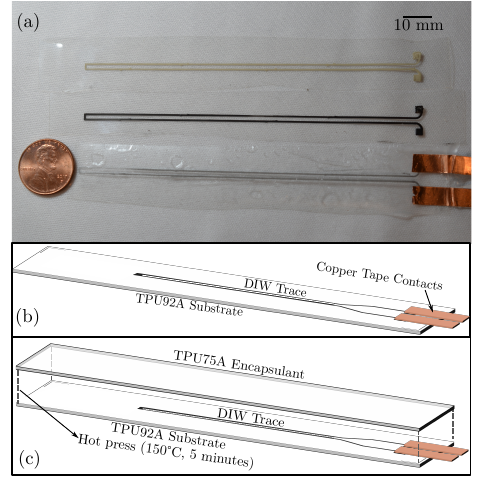}
    \caption{\revision{(a)} Stretchable strain sensors manufactured
    with different conductive materials: (Top)
    Silver-based sensor; (Middle) Carbon-based
    sensor; (Bottom) Gallium-Indium (ELMNT\textsuperscript{\textregistered}) sensor. \revision{Sensor dimensions are \SI{112}{\milli\metre} (length), \SI{2.2}{\milli\metre} (width), with a trace
    width of \SI{0.6}{\milli\metre}.
    }
    For further specifications on these materials,
    refer to Table~\ref{tab:1}.
    \revision{(b) Schematic showing the structure of conductor-elastomer composite based sensors. (c) Schematic for liquid metal based sensors. }}
    \label{fig:sensors}
\end{figure}

Figure~\ref{fig:sensors} shows three stretchable
strain sensors printed for this study, each using a different
conductive ink.
The suitability of each ink for the application at hand was
investigated 
via cyclic loading experiments in which
the sensors were repeatedly stretched as if they were
attached to the body of a 
continuum robot.
The sensors were evaluated in terms of
linearity, drift, and gauge 
factor. 
\subsection{Sensor Fabrication}
The sensors were fabricated via direct ink writing using
the V-One printer
(Voltera Inc., Waterloo, ON, Canada), equipped
with a \SI{0.23}{\milli\metre} diameter nozzle.
\revision{Print settings used in this work were: pass spacing of \SI{0.15}{\milli\metre}, dispensing height of \SI{0.08}{\milli\metre}, feed rate of \SI{500}{\milli\metre\per\min},
and trim length of \SI{25}{\milli\metre}.
This nozzle and these print settings result in a trace width of
\SI{0.3}{\milli\metre} per pass.}
Each sensor was printed onto a
thermoplastic polyurethane (TPU) film 
(ESTANE\textregistered~FS H92C4P, Lubrizol Corporation, Wickliffe, OH, USA), which 
serves as the base layer.
This specific TPU film was chosen because of its large
maximum recoverable strain (quoted to be 350\%),
its good adhesion to the three conductive inks
chosen for this study (which is introduced in the
next section),
and its higher rigidity compared to other
grades of TPU, which facilitates the handling
of the sensors after printing.
The sensor specimens used in this study were printed in batches of 4,
with a total manufacturing time of approximately 30 minutes per batch,
including the necessary post-processing time for each ink type specified below.
\revision{After completing the necessary post-processing, the sensors were attached to the desired surface using hot melt TPU film (ESTANE\textregistered~FS HM70A71, Lubrizol Corporation, Wickliffe, OH, USA) as an adhesive layer between the sensor substrate and the desired surface. The bonding was then completed by pressing the assembly in a \SI{1050}{\watt} hot press (TLM38385, Vevor, Rancho Cucamonga, CA, USA) at \SI{130}{\celsius} for 30 seconds to melt the TPU and ensure a robust bond. }
\subsection{Conductive Ink Types}
Table~\ref{tab:1} lists the three conductive inks
evaluated in this study and used in
the fabrication of the sensors shown in Fig.~\ref{fig:sensors}.
\revision{These inks were selected from a pool
of materials known to be well-suited for the 
fabrication of stretchable electronics~\cite{Fernandes2019}.
We narrowed down the pool by requiring that materials be commercially 
available, suitable for printing on TPU
substrates, and rated for
maximum elongation of, at least,
100\%.
Finally, we required materials to be compatible with direct
ink writing on the V-One printer.}
\begin{table}[]
\centering
\caption{Conductive Ink Types used for the Fabrication of the Stretchable Strain Sensors}
\begin{tabular}{@{}ccc@{}}
\toprule
\textbf{Name} & \textbf{Ink Type}                 & \textbf{Conductive Material} \\ \midrule
SE 1109       & Conductor-elastomer composite  & Silver                       \\
SE 1502       & Conductor-elastomer composite  & Carbon                       \\
ELMNT\textsuperscript{\textregistered} ST       & Liquid metal                      & Gallium-indium eutectic \\
\bottomrule
\end{tabular}
\label{tab:1}
\end{table}
While all the inks listed in Table~\ref{tab:1} are off-the-shelf and commercially available,
we note that none of them is specifically marketed for creating
strain sensors---instead, they are intended for printing conductive or 
resistive traces in electronic circuits;
therefore, the inks are being evaluated here for a
different application than the one they were originally 
intended for.
\subsubsection{Conductor-elastomer Composite Inks}
The SE 1109 and the SE 1502 (ACI Materials, Goleta, CA, USA) are elastomeric inks with conductive particles embedded into
them --- silver and carbon, respectively.
Sensors using these inks are cured, post-printing, by placing
them in an oven heated at \SI{140}{\celsius} for 5 minutes. 
The operation of stretchable strain sensors based on conductor-elastomer 
inks is explained by the
\textit{electrical percolation} phenomenon~\cite{Zhang2007,Flandin2001}, i.e., 
when the material is stretched, the
conductive particles embedded therein are pulled
apart from each other, producing an increase in 
electrical resistance. 
\subsubsection{Liquid Metal Inks}
The ELMNT\textsuperscript{\textregistered} ST (UES Inc., Dayton, OH, USA) is a
high-viscosity conductive ink based on a gallium-indium eutectic (eGaIn) alloy.
Since eGaIn is liquid at room temperature, fabricating
liquid metal sensors requires an additional step\revision{, as illustrated in Fig.~\ref{fig:sensors}(c)}:
A second TPU film 
(ESTANE\textregistered~FS  L75A4P, Lubrizol Corporation, Wickliffe, OH, USA) 
is hot pressed \revision{(\SI{150}{\celsius}, 5 minutes)} on top of the printed sensor to form an airtight seal with
the sensor’s bottom TPU
layer, encapsulating the liquid metal and preventing its escape.
Furthermore, liquid metal sensors require a
mechanical activation step, which involves either peeling off the TPU backing sheet or stretching the sensor \revision{to break the oxide shells on the liquid metal particles. Incomplete or imperfect
activation can result in a higher electrical resistance during the initial cycles of operation}.
For stretchable strain sensors utilizing liquid metal
inks, the increase in resistance is attributed
to the increase in length and decrease
in the cross-sectional area of the conductive pathway.
\subsection{Experimental Evaluation of the Conductive Inks}
\begin{figure*}
    \centering
    \includegraphics[width=0.75\linewidth]{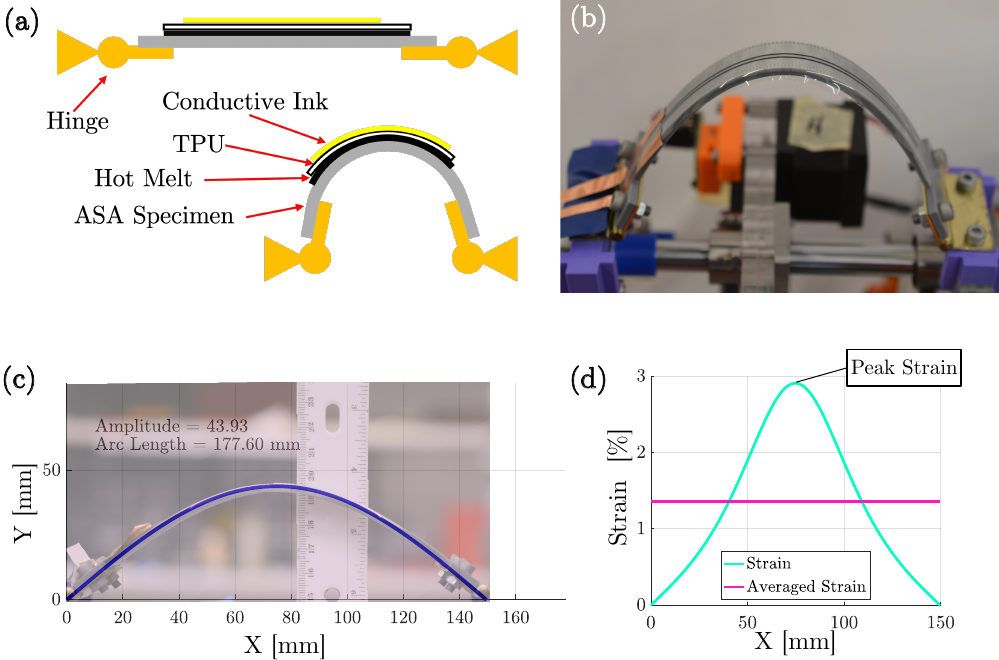}
    \caption{Experimental apparatus and test
    for the evaluation of the conductive inks.
    (a) Schematic of the test setup. Each sensor is attached to an Acrylonitrile Styrene Acrylate (ASA) beam, which is 
    controllably bent to simulate different
    loading conditions.
    (b) Actual ASA specimen under load with a
    sensor attached. Bending is achieved 
    by attaching one of the two ends of the
    beam to a linear stage controlled by
    a stepper motor.
    The electrical resistance of the sensor
    is monitored with an IM3536 LCR meter
    (Hioki E.E. Corporation, Nagano, Japan),
    not pictured here.
    The position of the moving terminal of the linear stage was
    recorded using a magnetic linear encoder, the
    AS5048A-HTSP (ams OSRAM AG, Premstätten, Austria), 
    and used to estimate the
    applied strain, as explained in the
    following.
    Both strain and resistance measurements were
    captured at \SI{10}{\hertz} and timestamped with
    Robot Operating System 2 (ROS 2) running on 
    Ubuntu 20.04, on a laptop equipped with an
    i7-8750H CPU (Intel Corp., Santa Clara, CA, USA).
    (c) Beam shape reconstruction based
    on the applied load, using a
    simple model of beam buckling under compressive loads~\cite{Hibbeler2023}.
    (d) Estimated strain distribution
    along the beam.
}
    \label{fig:hysteresis_exp}
\end{figure*}

The sensors were evaluated as illustrated in Fig.~\ref{fig:hysteresis_exp}:
Each sensor was first attached with hot melt TPU to a flat, 
\SI{3}{\milli\metre}-thick beam made of 
Acrylonitrile Styrene Acrylate (ASA), which was 
then cyclically bent from straight to maximum deformation and then back to straight 50 times by means of a motorized stage.
Three sensor specimens were prepared for each of the 
three inks, and their response 
was studied under four different experimental
conditions, detailed in Table~\ref{tab:sensor-characterization-experiment} and
labeled (a-d), aimed to
simulate different loading conditions that
the sensors could experience in practical use.
\begin{table}[]
\centering
\caption{Experimental Conditions used in the Evaluation of the Conductive Inks 
}
\begin{tabular}{@{}ccccc@{}}
\toprule
\textbf{ID} &
\multicolumn{1}{c}{\textbf{\begin{tabular}[c]{@{}c@{}}Strain\\ Rate (s$^{-1}$)\end{tabular}}} & \multicolumn{1}{c}{\textbf{\begin{tabular}[c]{@{}c@{}}Dwell\\ Time (s)\end{tabular}}} & \multicolumn{1}{c}{\textbf{\begin{tabular}[c]{@{}c@{}}Peak\\ Strain (\%)\end{tabular}}} & \multicolumn{1}{c}{\textbf{\begin{tabular}[c]{@{}c@{}}Average\\ Strain (\%)\end{tabular}}} \\ \midrule
(a) & 0.027                                                                                   & 1                                                                                     & 2.91                                                                                    & 1.36                                                                                       \\
(b) & 0.054                                                                                   & 5                                                                                     & 2.91                                                                                    & 1.36                                                                                       \\
(c) & 0.026                                                                                   & 5                                                                                     & 9.66                                                                                    & 2.63                                                                                       \\
(d) & 0.052                                                                                   & 1                                                                                     & 9.66                                                                                    & 2.63                                                                                       \\ \bottomrule
\end{tabular}
\label{tab:sensor-characterization-experiment}
\end{table}
Experimental parameters include the 
strain at full bending, the strain rate, and the dwell time.
Briefly, the strain at full bending describes the maximum
deformation experienced by the sensor in each experiment.
This parameter was controlled by regulating the
displacement of the motorized stage, and thus the beam's
curvature. 
As can be observed from Fig.~\ref{fig:hysteresis_exp}(d),
the strain distribution experienced by the beams during
testing was predictable but not uniform, therefore
Table~\ref{tab:sensor-characterization-experiment}
reports both peak and average values; the average value
will be used in the characterization of the gauge
factor, as explained below.
The strain rate is defined as the ratio of the average strain to the time duration of each cycle. It quantifies the speed at which
deformation occurs. This parameter was controlled by
regulating the speed of the motorized stage that was bending the
test beams (refer to Fig.~\ref{fig:hysteresis_exp}).
Finally, dwell time refers to the waiting period between two consecutive
bending cycles. 
The sensors were evaluated in terms of three performance metrics, namely
gauge factor, drift, and linearity of the response.
\subsubsection{Gauge Factor}
The gauge factor (GF) is defined as 
\begin{equation}
    \text{GF} = \frac{\Delta R/R_0}{\epsilon}.
    \label{eq:GF}
\end{equation}
where $R_0$ is the baseline electrical
resistance of the sensor (i.e., the
resistance under zero strain), $\epsilon$ is
the applied strain, and $\Delta R$ is the 
observed change in resistance.
For the purpose of this study, we estimate the 
GF by inputting the 
resistance change observed at full bending,
and the average strain created along the body of the sensor,
as listed in Table~\ref{tab:sensor-characterization-experiment}.
In general, a large GF is desirable,
as it corresponds to a more responsive
sensor.
\subsubsection{Drift}
Drift is a measure of how much the
sensor's baseline electrical resistance $R_0$ 
changes over time.
In stretchable strain sensors, drift is
often linked to the progressive accumulation of 
plastic deformation in the sensor's material,
which occurs as a result of repeated loading
cycles at high strain levels.
For the purpose of this study, 
we estimate drift as
\begin{equation}
     \text{Drift} = \frac{\Delta R_0/R_0}{\text{cycle}}.
    \label{eq:GF}
\end{equation}
We wish to identify an ink that
exhibits as little drift as possible, and
exhibits a response to strain that significantly
exceeds any drift that occurs within the
operational window.
\subsubsection{Linearity}
Linearity is simply estimated by finding 
the line that best approximates the sensor's 
response and calculating the corresponding
coefficient of determination ($\text{R}^\text{2}$).
High linearity is generally desirable as it
enables the simplest possible sensor calibration.
\subsection{Results}
\begin{figure*}[]
    \centering
    \includegraphics[width=0.8\linewidth, trim=-0 0 0 0,clip]{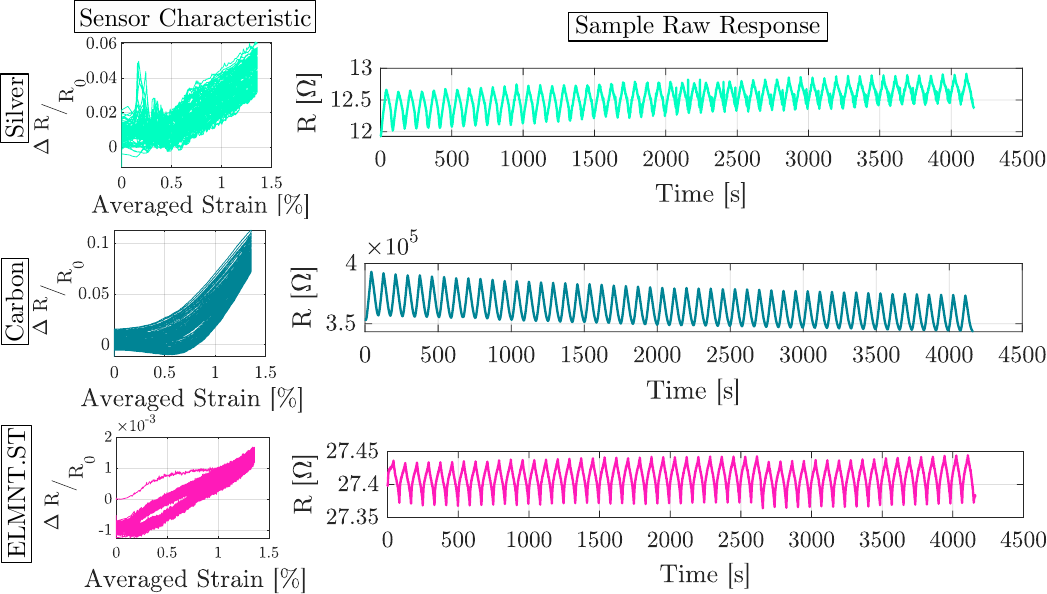}
    \caption{Raw sensor output and characteristic curves observed under experimental condition (a).
    The sensor using the ELMNT\textsuperscript{\textregistered} ink exhibited higher linearity 
    and significantly less drift than the ones based on the conductor-elastomer composite inks (silver and carbon). The initial resistance of the ELMNT\textsuperscript{\textregistered} sensor is slightly higher during the first cycle, attributed to imperfect mechanical activation.
    }
    \label{fig:sample-characterization-results}
\end{figure*}

The results of the ink evaluation experiments are summarized 
in Table~\ref{tab:hysteresis}, which reports the average 
values observed for each experimental condition.
\begin{table*}
\caption{Gauge Factor, Drift, and Linearity of the Proposed Sensors}
\centering
\begin{threeparttable}
\begin{tabular}{cccccc}
   \toprule
   Ink & Experimental Condition & GF & Drift ($\frac{\Delta R_0/R_0}{\text{cycle}}$) & Cycles\textsuperscript{\textdagger} & Linearity ($R^2$)\\
   \midrule
    SE 1109 (Silver) & (a)  & 3.0 & \SI{9.7e-4}{} & 43 & 0.98 \\
    & (b) & 3.5 & \SI{1.3e-3}{} & 37 & 0.98 \\
    & (c) & 15 & \SI{2.1e-2}{} & 19 & 0.93 \\
    & (d) & 14 & \SI{1.2e-2}{} & 30 & 0.90 \\
    \midrule
    SE 1502 (Carbon) & (a) & 5.5 & \SI{-3.4e-4}{} & \SI{2.2e2}{} & 0.93 \\
    & (b) & 5.2 & \SI{-4.5e-4}{} & \SI{1.6e2}{} & 0.95 \\
    & (c) & 2.0 & \SI{1.9e-4}{} & \SI{3.3e3}{} & 0.60\\
    & (d) & 1.8 & \SI{-2.0e-5}{} & \SI{2.4e3}{} & 0.60\\
    \midrule
    ELMNT\textsuperscript{\textregistered} ST & (a) & 1.2 & \SI{1.0e-5}{} & \SI{1.6e3}{} & 0.98 \\
    & (b) & 1.3 & \SI{2.0e-5}{} & \SI{9.0e2}{} & 0.99 \\
    & (c) & 0.88 & \SI{1.1e-4}{} & \SI{2.1e2}{} & 0.99 \\
    & (d) & 0.91  & \SI{1.0e-5}{} & \SI{2.4e3}{} & 0.99\\
   \bottomrule
\end{tabular}
\begin{tablenotes}
\footnotesize
\item \textsuperscript{\textdagger}: Number of cycles after which
the effect of drift becomes larger than the resistance change at full bending.
\end{tablenotes}   
\end{threeparttable}
\label{tab:hysteresis}
\end{table*}
The conductor-elastomer composite inks (silver and carbon) generally
exhibited larger but highly variable GF values. The silver-based sensors
showed approximately a 5-fold change, while the carbon-based sensors
exhibited about a 2.5-fold change over the tested ranges.
Both sensors were significantly affected by drift.
The plots in Fig.~\ref{fig:sample-characterization-results}
shows the sensors' responses under experimental condition (a) 
and visually illustrate the effect of drift on the sensor output.
In practical use, these sensors would need frequent recalibration to compensate for drift,
potentially as often as every 19 bending cycles for the silver-based sensor (see Table~\ref{tab:hysteresis}). Interestingly,
the carbon-based sensors exhibited significantly less drift when subjected to high strain
(experimental conditions (c) and (d)), but also significant degradation
in linearity under these conditions, with the $R^2$ score dropping to $\approx$ 0.6.
In contrast, the liquid metal sensors exhibited GFs ranging between
0.88 and 1.3 and
demonstrated significantly better stability, with at most $\sim1.5\times$ variation over the same tested ranges. Furthermore, these sensors exhibited 
high linearity across all 
experimental conditions ($R^2 > 0.98$) and 
significantly less drift than the conductor-elastomer sensors.

\subsection{Discussion}
\revision{
Numerous studies have investigated materials for fabricating
stretchable strain sensors, as summarized in a recent survey
by Zhao \textit{et al.}~\cite{Zhao2025}.
Many of these studies focus on experimental nanomaterials
that are typically available only at a laboratory scale.
In contrast, in this work, we exclusively consider
commercially available inks.}
\revision{Carbon and silver elastomeric inks are composed of
networks of conductive fillers embedded within an
elastomeric matrix. While these networks can produce
relatively high gauge factors (GF) when stretched,
they are prone to damage and permanent deformation
resulting in drift in electrical resistance.
The elastomeric sensors discussed in this paper
exhibit behavior similar to those described in
the literature. For instance, Zhang and
colleagues~\cite{Zhang2018} studied stretchable
strain sensors made of carbon and silver particles
within a TPU matrix.
These sensors achieved a GF ranging from 5 to 10
for strains up to 10\% and exhibited drift in the
order of 1 $\times$ 10\textsuperscript{-4} per cycle
under repeated stretching at the same strain level.
The GF of these sensors is comparable to that of
the silver and carbon sensors tested in this study,
while the drift results align with those observed
for the carbon sensor but are smaller than those
seen in the silver sensor.}
\revision{
In contrast to carbon and silver inks,
ELMNT\textsuperscript{\textregistered} ink is liquid at
room temperature. The only significant drift affecting
sensor performance stems from the TPU material
on which the ELMNT\textsuperscript{\textregistered} ink
is printed and encapsulated, leading to minimized drift
and more stable sensor behavior.
The liquid metal sensors examined in this study can be
compared to those described by Abbara and
colleagues~\cite{Abbara2023}, who used a syringe-dispensing
process to print ELMNT\textsuperscript{\textregistered}
liquid metal traces onto TPU substrates. The sensors
in~\cite{Abbara2023} use larger trace width
(\SI{1}{\milli\metre}, compared to \SI{0.6}{\milli\metre}
in this study) and were tested under
constant strain cycling while varying temperature.
These sensors demonstrated a GF of 1.2 to 1.4 (average 1.3)
and exhibited negligible drift at room temperature.
Similarly, the liquid metal sensors reported here
achieved GFs of 1.2 to 1.3 under low peak strain
conditions (2.91\% peak strain in experimental
conditions (a) and (b)), consistent with findings
in~\cite{Abbara2023}. However, under higher peak
strain conditions (9.66\% peak strain in
experimental conditions (c) and (d)), the GF was
slightly lower, at approximately 0.9. The variation
in GF across strain ranges may be attributed to
differences in TPU substrates or other, as yet
unidentified, factors warranting further
investigation. Notably, the sensors studied here
maintained extremely low drift, consistent with the
findings of~\cite{Abbara2023}.}

\revision{Although the ELMNT\textsuperscript{\textregistered}
liquid metal sensors exhibited the lowest GF among the
tested sensors, they proved to be the most suitable
for shape sensing applications due to their negligible
drift, which enables a clear and consistent sensor
response. Additionally, their relatively
uniform GF across varying experimental conditions,
along with high sensor linearity, are important benefits.}
In the next section, we describe experimental work
aimed to validate the viability of using
liquid metal sensors
for the shape sensing of continuum robots.
% %%

\section{Shape Sensing Experiments}
\begin{figure}
    \centering
    \includegraphics[width=0.7\linewidth, trim=-0 0 0 0,clip]{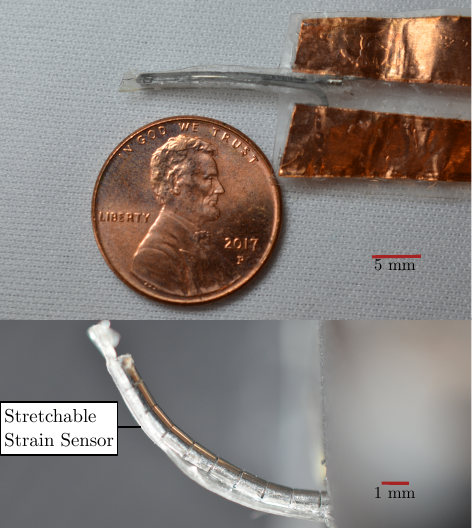}
    \caption{(Top): ELMNT\textsuperscript{\textregistered} stretchable strain sensors
    installed onto a notched tube wrist~\cite{Pacheco2021}.
    \revision{The sensing area measures \SI{17}{\milli\metre} in length and \SI{0.6}{\milli\metre} in width, with a trace width of \SI{0.3}{\milli\metre}.}}
%\PR{change to 0.3mm}    }
    \label{fig:NWT_dimension}
\end{figure}

\begin{figure}
    \centering
    \includegraphics[width=0.7\linewidth, trim=-0 0 0 0,clip]{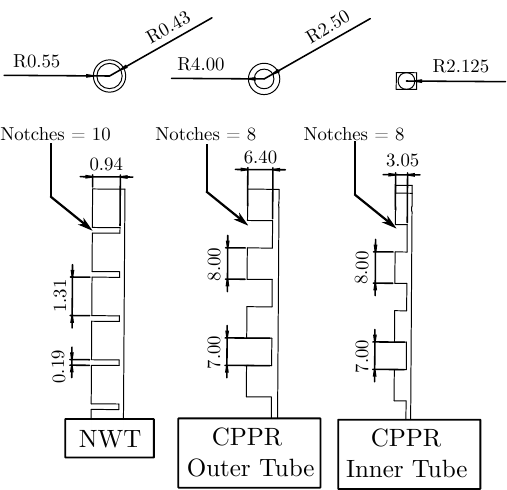}
    \caption{\revision{Dimensions of the continuum robots utilized in this study. All figures are in millimeters, except for the notch count.}
    }
    \label{fig:CR-parameters}
\end{figure}

To study the suitability of liquid metal stretchable strain sensors
for shape sensing, we performed experiments with 
two types of continuum robots, namely a 
concentric push-pull robot (CPPR)~\cite{Oliver2021}
and a notched tube
wrist (NWT)~\cite{Pacheco2021}.
These robots are shown in Fig.~\ref{fig:fig-1} and
Fig.~\ref{fig:NWT_dimension}, respectively.
Details on the actuation and the modeling of these
devices are beyond the scope of this manuscript,
and interested readers are referred to the
papers cited above.
The body of the CPPR is made of ASA, and it 
was printed with a Bambu Lab 
X1 printer (Bambu Lab, Shenzen, PRC). The outer
diameter of this robot is \SI{8}{\milli\metre}.
The NWT was manufactured by laser-cutting notches
in the body of a thin Nickel-Titanium tube of
diameter \SI{1.1}{\milli\metre}.
Laser cutting was outsourced to Resonetics
LLC (San Diego, CA, USA).
Other robot dimensions are listed in Fig.~\ref{fig:CR-parameters}.
\subsection{Procedure}
The two manipulators were outfitted with a liquid metal
stretchable strain sensor,
which was attached to the outer surface with hot melt
TPU as illustrated in Fig.~\ref{fig:fig-1} and
Fig.~\ref{fig:NWT_dimension}, and were subsequently subjected to a cyclic loading 
experiment wherein they were bent up to their tightest bending
radius.
Each experiment was repeated three times, during which 
the sensor response was logged together with the 
manipulator bending angle.
\revision{The bending angle of the robots was measured via image processing, using video footage
captured with a RealSense D405 camera (Intel Corp., Santa Clara, CA, USA) at a resolution of 640 × 480 pixels.
Both manipulators were outfitted with 
colored markers (visible in Fig.~\ref{fig:fig-1}) in order to facilitate the detection 
of their tips during bending.}
\begin{figure*}
    \centering
    \includegraphics[width=1\linewidth, trim=-0 0 0 0,clip]{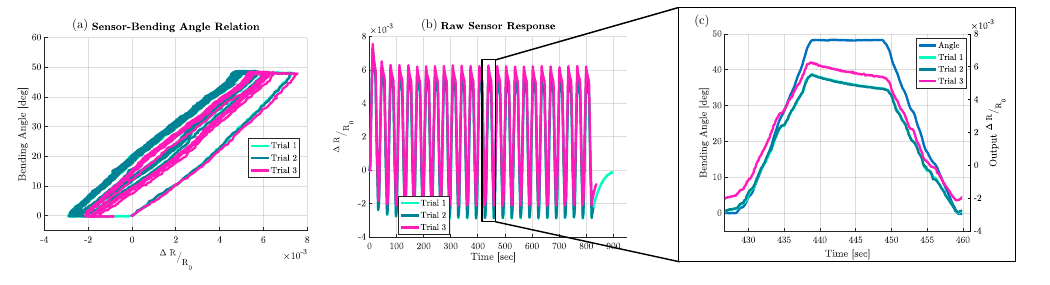}
    \caption{CPPR cyclic loading experiment. (a):  Sensor response to bending angle. (b): Raw sensor output. (c): Enlarged view of the sensor's response to CPPR bending.} 
    \label{fig:CAAR_results}
\end{figure*}
\begin{figure*}
    \centering
    \includegraphics[width=1\linewidth, trim=-0 0 0 0,clip]{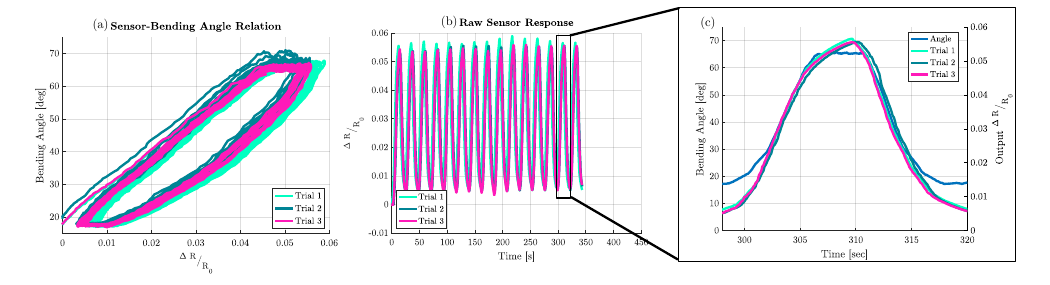}
    \caption{NWT cyclic loading experiment. (a): Sensor response to bending angle. (b): Raw sensor output. (c): Enlarged view of the sensor's response to NWT bending.}
    \label{fig:NWT_results}
\end{figure*}

\subsection{Results}
Results are shown in Fig.~\ref{fig:CAAR_results} and Fig.~\ref{fig:NWT_results} for the CPPR and the 
NWT, respectively.
In these figures, sub-figure (a) displays
resistive response to bending angle, (b) presents raw response of the sensor, and (c) illustrates the single-cycle 
relationship between the sensor output and 
the manipulator's bending angle.

\section{Discussion}
Experimental findings suggest the viability of liquid metal stretchable
strain sensors for the shape sensing of continuum
robots.
Experimental results reported in
Figs.~\ref{fig:CAAR_results} and~\ref{fig:NWT_results}
show generally good linearity and repeatability, with
some caveats as discussed in the following.
It is important to note that the strain encountered 
by the sensors in these experiments was larger than
the average strain encountered in earlier testing.
In the continuum manipulators used for this study,
the maximum strain is experienced on the outer surface
of the robot's body. 
Using the relations from~\cite{swaney_design_2017},
we estimate the CPPR to have experienced a maximum
strain of 3.15\% at full bending, while the NWT
experienced a maximum strain of 7.40\%.
These findings indicate that the sensors can
effectively handle the strains experienced by
continuum robots.
The data in Figs.~\ref{fig:CAAR_results}
and~\ref{fig:NWT_results} also reveal some undesirable
effects. First, in sub-figures (a), there is noticeable hysteresis in the
sensor response, particularly in the case of the experiment with the NWT.
\revision{These results highlight
the need for hysteresis compensation 
in order to achieve accurate sensor
calibration.}
\revision{In a related study~\cite{Chen2013}, a stretchable strain sensor—fabricated by mixing carbon black particles into liquid polyurethane rubber, followed by curing and cutting into strips—was employed for sensing and closed-loop control of a continuum robot. Although that sensor exhibited significant viscoelasticity, hysteresis, and creep, its behavior was effectively modeled, enabling precise control. In future work, we plan to adopt a similar approach by implementing mathematical compensation models to address hysteresis. }

\revision{Sensor response on the CPPR 
exhibited drift during the initial
bending cycles, as can be observed in
Fig.~\ref{fig:CAAR_results}(b).
We attribute such drift to an incomplete
activation of the liquid
metal ink prior to the experiment.
As it was discussed earlier in the
manuscript, each sensor needs to be 
\textit{broken in} before use by means of
repeated stretching. The purpose of this
process is to fully break the insulating
gallium oxide shell around the gallium
indium liquid metal particles in the ink,
allowing the liquid metal to form a
continuous conductor.
In cases of imperfect activation, some
liquid metal particles retain their oxide
shell. 
Then, during the first few cycles of operation, 
some of these remaining oxide
shells break, resulting in the release of
the liquid metal, which in turn leads to a
decrease in resistance.}
\revision{An important limitation of the sensors
described in this manuscript is their ability to
measure only the total strain of a continuum robotic
arm as a whole.
These sensors perform well for robots bending in a
constant curvature arc but are not suitable for robots
that deform into more complex shapes.
To address this limitation, future work will focus on
extending the current approach to enable local strain
measurements.
This could be achieved, for instance, by fabricating
an array of miniaturized strain gauges distributed along
the robot's body.
Although this solution is not explored in the present
manuscript, we believe that our sensor fabrication
method holds promise for producing such miniaturized
sensors, thanks to the ability to print traces
as small as \SI{0.3}{\milli\metre}}.

\section{Conclusion}
This paper presented the design, fabrication, and
experimental characterization of stretchable strain sensors
for the shape sensing of continuum robots.
\revision{Among the different sensor materials evaluated
in the study, gallium-indium eutectic emerged as the
most suitable for the application at hand,
enabling the creation of sensors with
minimal drift and high linearity.
Experimental findings demonstrate the viability of
the proposed sensors for shape sensing at both
centimeter and millimeter scales.}
\revision{Future directions for this work include
sensor calibration, focusing on modeling the 
observed hysteresis and ``break-in'' behavior.
Additionally, this proof-of-principle study was
limited to unidirectional bending. In practical
applications, multiple sensors would be required to
monitor multiple sections of a continuum robot,
enabling local shape sensing under complex bending
conditions.
Exploring these scenarios represents another key area
of focus for our future work.}
%%

% \newpage
\bibliographystyle{IEEEtran}
\bibliography{DIW_Sensor}

\end{document}